\documentclass[sigconf]{acmart}

\AtBeginDocument{%
  \providecommand\BibTeX{{%
    \normalfont B\kern-0.5em{\scshape i\kern-0.25em b}\kern-0.8em\TeX}}}

\copyrightyear{2020}
\acmYear{2020}
\setcopyright{acmcopyright}\acmConference[MM '20]{Proceedings of the 28th ACM International Conference on Multimedia}{October 12--16, 2020}{Seattle, WA, USA}
\acmBooktitle{Proceedings of the 28th ACM International Conference on Multimedia (MM '20), October 12--16, 2020, Seattle, WA, USA}
\acmPrice{15.00}
\acmDOI{10.1145/3394171.3413995}
\acmISBN{978-1-4503-7988-5/20/10}

\acmSubmissionID{700}

\usepackage{booktabs} 
\usepackage{subfigure}
\usepackage{multirow}
\usepackage{comment}
\usepackage[american]{babel} 
\usepackage{microtype} 
\usepackage{tikz}
\usepackage{bbm}
\usepackage{amsmath}
\newcommand{\D}{\mathcal{D}} 
\newcommand{\X}{\mathcal{X}} 
\newcommand{\Y}{\mathcal{Y}} 
\newcommand{\x}{\boldsymbol{x}} 
\newcommand{\R}{\mathbb{R}}
\newcommand{\p}{\boldsymbol{p}}
\newcommand{\q}{\boldsymbol{q}}

\newcommand{\f}{\boldsymbol{f}} 

\begin{document}
\fancyhead{}

\title{Simultaneous Semantic Alignment Network for \\Heterogeneous Domain Adaptation}

\author{Shuang Li$^{1}$, Binhui Xie$^{1}$, Jiashu Wu$^{2}$, Ying Zhao$^{1}$, Chi Harold Liu$^{1,*}$, Zhengming Ding$^{3}$}

\thanks{$^\ast$Corresponding author.}
\affiliation{%
  \institution{$^1$School of Computer Science and Technology, Beijing Institute of Technology, Beijing, China.}
  \institution{$^2$School of Computing and Information Systems, The University of Melbourne, Melbourne, Australia.}
  \institution{$^3$Department of Computer, Information and Technology, Indiana University-Purdue University Indianapolis.}
}
\email{{shuangli, binhuixie, yingzhao, chiliu}@bit.edu.cn, jiashu@student.unimelb.edu.au, zd2@iu.edu}

\renewcommand{\shortauthors}{Li et al.}

\begin{abstract}
    Heterogeneous domain adaptation (HDA) transfers knowledge across source and target domains that present heterogeneities e.g., distinct domain distributions and difference in feature type or dimension. Most previous HDA methods tackle this problem through learning a domain-invariant feature subspace to reduce the discrepancy between domains. However, the intrinsic semantic properties contained in data are under-explored in such alignment strategy, which is also indispensable to achieve promising adaptability. In this paper, we propose a Simultaneous Semantic Alignment Network (SSAN) to 
    simultaneously exploit correlations among categories and align the centroids for each category across domains. In particular, we propose an implicit semantic correlation loss to transfer the correlation knowledge of source categorical prediction distributions to target domain. 
    Meanwhile, by leveraging target pseudo-labels, a robust triplet-centroid alignment mechanism is explicitly applied to align feature representations for each category. Notably, a pseudo-label refinement procedure with geometric similarity involved is introduced to enhance the target pseudo-label assignment accuracy.
    Comprehensive experiments on various HDA tasks across text-to-image, image-to-image and text-to-text successfully validate the superiority of our SSAN against state-of-the-art HDA methods. The code is publicly available at https://github.com/BIT-DA/SSAN.
\end{abstract}




\maketitle

\vspace{-2mm}
\begin{figure}[!htbp]
    \centering
    \includegraphics[width=0.48\textwidth]{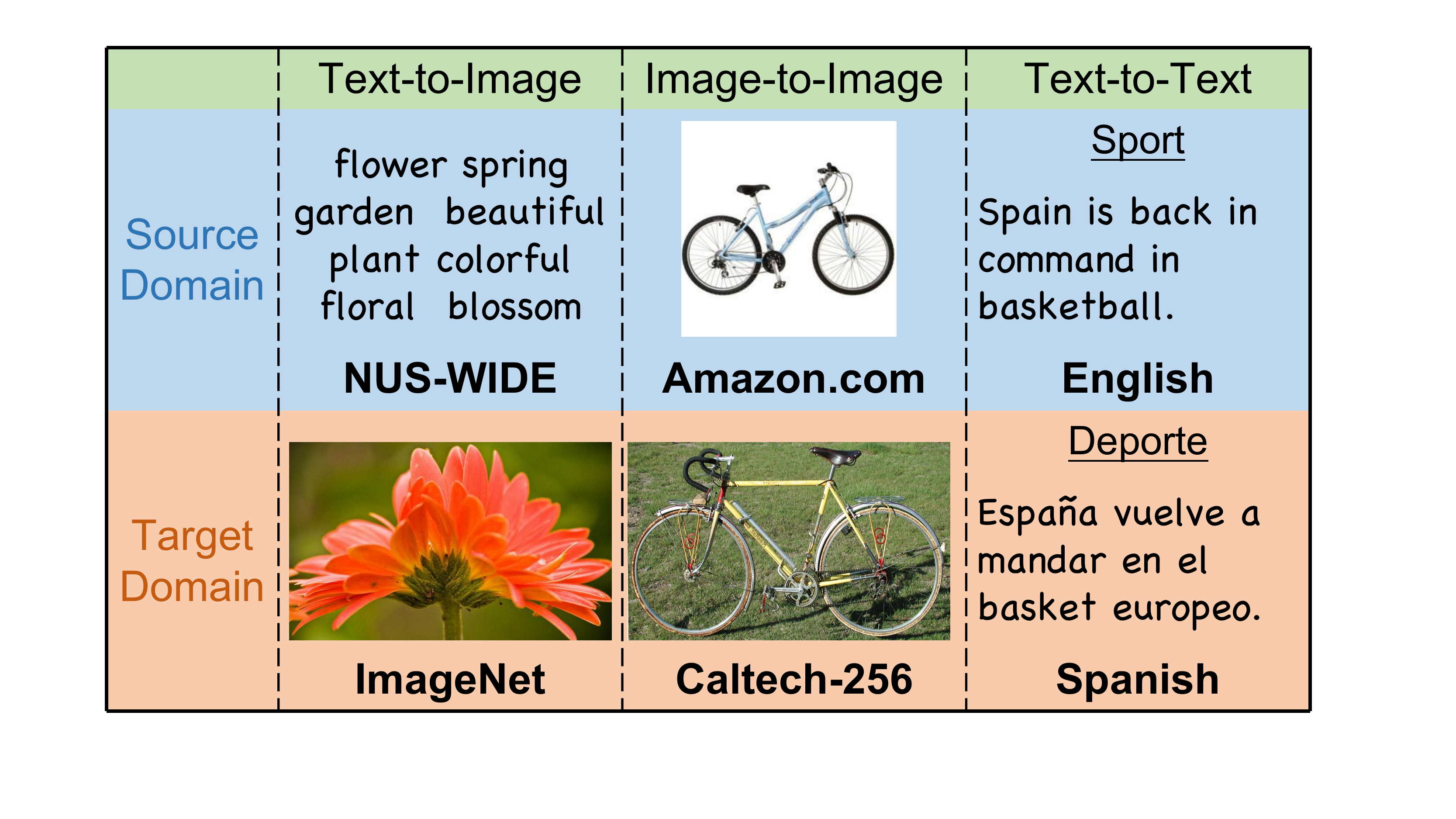}\vspace{-3mm}
    \caption{Examples of heterogeneous domain adaptation from left to right are 1) text-to-image classification with distinct data modalities; 2) object recognition with different feature dimensions; 3) cross-lingual document categorization.} 
    \label{Fig_HDA}
\end{figure}\vspace{-2mm}

\section{Introduction}
A vast majority of learning scenarios require sufficient amounts of labeled training data to achieve promising classification and generalization performance in a target domain. However, due to the expensive cost of data collection and laborious annotation for each problem of interest, it is crucial to transfer the enriched knowledge from an auxiliary domain (a.k.a., source domain) to facilitate learning a robust model for a target domain~\cite{survey}. Towards this goal, a plenty of domain adaptation (DA) techniques have been successfully applied in various multimedia tasks such as multimodal learning~\cite{multimodal_learning1,multimodal_learning2,DTN}, visual object recognition~\cite{visual_object_recognition,snell2017prototypical,DCAN}, and text categorization~\cite{text_categorization1,text_categorization3,text_categorization2}. It is worth noting that most existing DA efforts are based on the assumption that the data from different domains are represented by the same type of features~\cite{luo2017nips,bendavid2006analysis,DAN-PAMI,ADDA,DANN,ICCV_cite,DRCN}. In other words, they cannot cope with the situation where the source and target domain data do not share the same feature representations, which is frequently encountered in real-world applications. Figure~\ref{Fig_HDA} presents three examples that involve data from different feature spaces. Take cross-lingual document categorization as an example, documents in English will not share the same feature space with those in Spanish, owing to distinct vocabularies varied among languages.

Heterogeneous domain adaptation (HDA) attempts to address the task of associating cross-domain data observed in separate feature spaces, which vary greatly from one dataset or domain to another~\cite{MMDT_ICLR, HFA}. In the literature, most current approaches focus primarily on learning one or two feature transformations to eliminate the heterogeneities in the data and hence facilitate the divergence minimization between domains. Formally, one can map data from one domain to the other~\cite{ARC-t,MMDT_IJCV,SSKMDA,LS-UP,MAPHERE} or map data into a common feature subspace~\cite{HeMAP,DTN,HTDCC,SCP,SHFA}. For instance, Kulis et al.~\cite{ARC-t} first proposed an asymmetric regularized cross-domain transformation (ARC-t) to learn asymmetric and non-linear transformations with label information guarantees. While Shi et al.~\cite{HeMAP} presented a heterogeneous spectral mapping (HeMap), which derives two projection matrices based on spectral embedding.

Despite the great success achieved by these HDA methods, they would suffer from a major limitation: the instances of different categories are confounded together during the brute-force alignment process. As a result, they fail to produce discriminative features so that how to appropriately leverage the semantic properties underlying the data remains a significant issue. On one hand, the semantic correlations between classes have not been uncovered. The correlation relationship such as \emph{laptop} should be more semantically similar to \emph{monitor} than \emph{mug} is useful to guide better alignment between domains and hence worth excavating. On the other hand, even if semantic alignment for traditional domain adaptation has received increasing attention~\cite{MSTN,SDTADT}, to our best knowledge, 
semantic alignment has not been applied to address HDA problems due to separate feature spaces between domains, which becomes even harder to tackle when the labeled target data are limited.

To overcome the aforementioned challenges, we take a different tact and present a \textit{Simultaneous Semantic Alignment Network (SSAN)} for HDA problems. In this work, we pursue how to contemporaneously transfer the semantic correlation knowledge of the original data as much as possible and yield discriminative and domain-invariant feature transformations. More precisely, to deal with the data 
that present heterogeneities, we first build two non-linear feature encoders using neural networks (one for each domain), which allow us to generate a common feature subspace. This common feature subspace can be used to carry out diverse features with arbitrary dimensionality. Hereafter, we utilize all labeled source data to train a shared classifier with the standard supervised classification loss. Meanwhile, with the observation that the classifier shares the similar prediction distribution for the same class between the labeled source and target data, we propose an \emph{implicit semantic correlation} loss, which extends knowledge distillation~\cite{model_compression, hinton2015distilling} to the HDA scenario. Thus, we can achieve alignment of the semantic correlations among categories across domains. 

However, when optimizing the above two objectives to derive discriminative structures of supervised data in both domains, the unlabeled target data are failed to be fully excavated. As noted in~\cite{SHFA}, exploring unlabeled target instances during training would be helpful for HDA problems. To this end, we first strive to assign pseudo-labels for unlabeled target instances by ensemble predictions that reach consensus by both neural networks (the shared classifier) and the geometric similarity mechanism. In light of this, SSAN will automatically strengthen the confidence of pseudo-labels by using this pseudo-label refinement process. Furthermore, to reduce the bias caused by false pseudo-labels, we design a triplet-centroid alignment mechanism to match the centroids of transformed source, target and source-target combined features in each category, named as \emph{explicit semantic alignment}. Ultimately, for the purpose of more stable adaptation, we also consider a simple yet efficient way to facilitate distribution alignment with a domain confusion loss.

In summary, the contributions of our work are highlighted below:
\vspace{-4mm}
\begin{itemize}
    \item To our best knowledge, this is the first time to successfully utilize the semantic properties within data to solve HDA problems, promoting better adaptation capability.
    \item We present a novel Simultaneous Semantic Alignment Network, named SSAN, to consider and excavate the discriminative semantic structures of distinct distributions in HDA.
    \item Comprehensive evaluations on transfer tasks of text-to-image, image-to-image and text-to-text demonstrate the competence of our model, exceeding state-of-the-art HDA approaches. 
\end{itemize}

\section{Related Work}

Leveraging the knowledge extracted from the source domain, traditional domain adaptation~\cite{CDAN,JADA,TCA,DICD,DUCDA,PFAN} aims to tackle the task of associating the homogeneous data described by the same type of features across distinct domains. Another branch of works follows the heterogeneous setting, where the data are not only drawn from dissimilar distributions but also represented by features with different dimensions or modalities, making it challenging to leverage knowledge obtained from source data to assist the target learning task. Here, we delve into heterogeneous domain adaptation works. In recent years, heterogeneous domain adaptation (HDA) has attracted increasing attention~\cite{STN,MMDT_IJCV,SHFA,PA}. The proposed SSAN distinguishes from existing HDA methods in two aspects: 

\textbf{Implicit Semantic Correlation Transfer}. When bridging the gaps across domains, existing methods only leverage information extracted from the feature-level. Wang and Mahadevan~\cite{DAMA} proposed a manifold alignment method (DAMA) to preserve label information during the alignment process. Duan et al.~\cite{HFA} presented Heterogeneous Feature Alignment (HFA), which incorporates augmented features and the standard SVM for performing recognition. Li et al.~\cite{SHFA} extended HFA to a semi-supervised version (SHFA) by utilizing unlabeled target data during training. To learn a domain-invariant representation via an asymmetric category-independent transform, Hoffman et al.~\cite{MMDT_ICLR} proposed Max-Margin Domain Transforms (MMDT). Zhou et al.~\cite{SHFR} presented Sparse Heterogeneous Feature Representation (SHFR) to cast the learning of feature mappings as a compression sensing problem. Yan et al.~\cite{SGW} put forward a semi-supervised entropic Gromov-Wasserstein discrepancy (SGW) approach to learn optimal transport from source to target domain features. Inspired by recent advances in deep learning, Chen et al.~\cite{TNT} proposed Transfer Neural Trees (TNT) with stochastic pruning to solve feature mapping and facilitate adaptation. 

However, none of these methods utilize the semantic correlations contained in the predictions which can effectively guide the alignment process and facilitate better transferability.

\begin{figure*}[!htbp]
    \centering
    \includegraphics[width=0.90\textwidth]{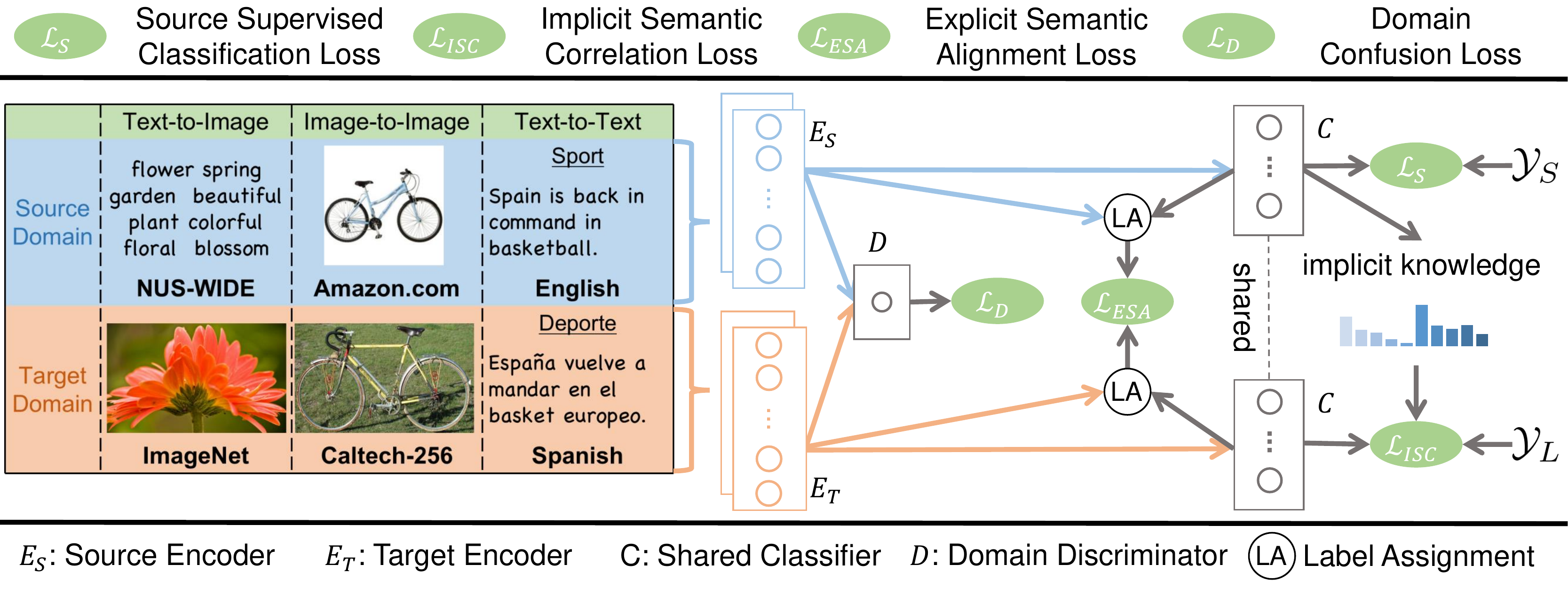}\vspace{-4mm}
    \caption{Approach overview. We employ an implicit semantic correlation loss over labeled target data to imitate the learned source prediction distributions. Simultaneously, we explicitly restrict the centroids of source, target and the combination of source and target in each class to be closer. To take advantage of the unlabeled target instances in the alignment process, we assign pseudo-labels by the consistent predictions between neural network (the shared classifier) and geometric similarity mechanism. Besides, we apply a domain confusion loss to learn domain-invariant feature representations.}
    \label{Fig_architecture}\vspace{-4mm}
\end{figure*}

\textbf{Explicit Semantic Alignment}. There exist several methods that apply pseudo-labels of unlabeled target instances to enforce the semantic alignment across domains. To learn representative cross-domain landmarks to derive proper feature subspace that eliminates the domain divergence, Tsai et al.~\cite{CDLS} presented Cross-Domain Landmark Selection (CDLS). Similarly, Hsieh et al.~\cite{GJDA} considered Generalized Joint Distribution Adaptation (G-JDA), which jointly matches both marginal and conditional distributions for adaptation and classification.  
All of this, unfortunately, are vulnerable to the error accumulation as the pseudo-label accuracy is not guaranteed, which is an important origin of performance reduction. Recently, to avert false pseudo-labels introduced by traditional hard label assignment, Yao et al.~\cite{STN} proposed a Soft Transfer Network (STN) to adopt a soft-label strategy during class-level alignment. 

In contrast, 
we introduce a geometric similarity pseudo-label refinement mechanism to automatically strengthen the confidence of pseudo-labels. With the objective of learning a more robust model, we further design a triplet-centroid alignment mechanism to match the centroids of transformed source, target and source-target combined features for each category in a progressive manner.

\section{The Proposed Algorithm}


\subsection{Preliminary and Motivation}

For heterogeneous domain adaptation (HDA) with a semi-supervised setting, we have one labeled source domain and one scarcely labeled target domain. To be concrete, let $\D_S=\{\X_S\,,\Y_S\}=\{(\x_{s_i}\,, y_{s_i})\}_{i=1}^{n_s}$ denote a set of training instances of the source domain, where $\x_{s_i} \in \R^{d_s}$ denotes the $i$-th example with $d_s$-dimensional features and $y_{s_i} \in \{1,\ldots,K\}$ represents the corresponding label. Similarly, let $\D_L=\{\X_L\,,\Y_L\}=\{(\x_{l_i}\,, y_{l_i})\}_{i=1}^{n_l}$ and $\D_U=\{\X_U\}=\{\x_{u_i}\}_{i=1}^{n_u}$ denote labeled and unlabeled instances of the target domain, respectively, where $\x_{l_i}\,,\x_{u_i} \in \R^{d_t}$ and $y_{l_i} \in \{1, \ldots, K\}$. Notably, $d_s \ne d_t$ and $n_l \ll n_u$. In a nutshell, our ultimate goal is to produce a shared classifier $C$ that can be applied to encoded features $\f(\x_i)$ of source and target data and correctly predict labels of the unlabeled target instances. For simplicity, we define that $\f_S(\x_i)=E_S(\x_i)$ if $\x_i\in\X_S$ and $\f_T(\x_i)=E_T(\x_i)$ if $\x_i\in\X_L\cup\X_U$, where $E_S$ and $E_T$ are source and target feature encoders, respectively.

Generally, most prior researches in HDA are devoted to associating heterogeneous data by learning a domain-invariant feature subspace.
However, a drawback of these methods is that they ignore the semantic information contained in instances. 
Even though some methods enforce the alignment between domains in  a brute-force manner, they may still distort the features with the same class label due to the strong heterogeneities present across domains, which will damage the discriminative structures. To alleviate this concern, we aim to build two feature encoders $E_S\,,E_T$ for source and target inputs, respectively, to yield a domain-invariant and discriminative feature subspace. Besides the standard supervision loss, we also explore the implicit knowledge learned from the labeled source data, transferring the correlations among categories. To mitigate the negative transfer caused by falsely-assigned pseudo-labels, we design a geometric pseudo-label refinement mechanism which can leverage the geometric property of the original data to boost pseudo-label assignment accuracy. 
Furthermore, to learn semantic representations for unlabeled target data, we propose the explicit semantic alignment in a progressive way by matching centroids corresponding to each category in the source, target and source-target combined feature representations via refined target pseudo-labels. 

\subsection{Implicit Semantic Correlation Transfer}

Deep learning has been introduced into HDA and achieves remarkable performance across multimodal transfer tasks~\cite{STN,TNT, DTN}. In general, it is the premise to train a discriminative classifier by minimizing the empirical error on labeled source data. Technically, we represent the source supervised classification loss as follows:
\begin{small}
    \begin{align}\label{eq-supervised-loss}
        \mathcal{L}_{S}=\mathcal{J}_{sup}(\X_S,\Y_S) = \frac{1}{n_s}\sum_{\x_i \in \X_S \,,y_i\in\Y_S} \mathcal{J}_{ce}\left(C(\f_S(\x_i)),y_i\right)\,,
    \end{align}
\end{small}where $\mathcal{J}_{ce}(\cdot,\cdot)$ is the cross-entropy loss. 
By the same token, we could utilize the limited target labeled data to learn the parameters of the target network (target encoder $E_T$ and the shared classifier $C$) according to Eq.~(\ref{eq-supervised-loss}). However, this often results in overfitting to the limited labeled target data, which suffers from performance degradation when predicting in the unlabeled target domain. 

In spite of the fact that the source and target domains present huge heterogeneity, it is reasonable that the classifier should produce the similar categorical probability distribution for the same category in both domains, which is referred to as implicit knowledge. This is intuitive, for instance, \emph{laptop} should have a closer semantic correlation with \emph{monitor} and \emph{keyboard} than \emph{bike} and \emph{mug} no matter whether it comes from the source domain or the target domain. Thus, we realize that the source category implicit knowledge could promote the generalization performance of classifier on the target domain. Therefore, we regard the average over the probabilistic outputs of source instances in class $k$ as the $k$-th teacher and denote it as $\q^{(k)} \in \R^K$, which is the ``soft label'' for class $k$. Owing to plenty of labeled source data available, ``soft labels'' contains much valuable category correlations. To adequately capitalize on these correlations, we adopt a softmax with a higher temperature $T$ to smooth the classifier activations, which could produce a softer probability distribution over classes. Thus, we define $\q^{(k)}$ as 
\begin{small}
   \begin{align}\label{eq-source-soft}
    \q^{(k)} = \frac{1}{\lvert \X_S^{(k)} \rvert}\sum_{\x_{i}\in\X_S^{(k)}} softmax\left(\frac{C\left(\f_S(\x_{i})\right)}{T} \right) \,,
   \end{align} 
\end{small}where $\X_S^{(k)}$ denotes the set of source instances of class $k$ and $\lvert \cdot \rvert$ represents the number of samples in the set. Given a labeled target instance, we could fine-tune the target network with ``soft labels'' to learn and transfer the semantic correlations from source to target domain. Hence, with the supervision of the learned ``soft labels'', the corresponding loss can be calculated as
\begin{small}
    \begin{align}\label{eq-soft-loss}
        \mathcal{J}_{soft}\left(\X_L, \Y_L\right) = -\frac{1}{n_l}\sum_{\x_{i} \in \X_L\,,y_i\in\Y_L} {\q^{(y_i)}}^\top \log \p_i\,,
    \end{align}
\end{small}where $\p_i$ is the probabilistic output for labeled target instance $\x_i$, and $\p_i=softmax\left(C(\f_T(\x_{i}))\right)$. The above loss can transfer the semantic correlations from the source network to the target network. To conclude, we further take the supervised loss of labeled target data into consideration and define the implicit semantic correlation loss as 
\begin{small}
    \begin{align}\label{eq-ISC-loss}
        \mathcal{L}_{ISC} = \left(1-\alpha\right) \mathcal{J}_{sup}\left(\X_L, \Y_L \right) + \alpha \mathcal{J}_{soft}\left(\X_L, \Y_L \right) \,.
    \end{align}
\end{small}In this way, the target network can perform better generalization to these instances around class boundaries and capture the semantic correlations among categories in supervised data, which will achieve significant performance gains. 

\subsection{Explicit Semantic Alignment}

Recall that we dedicate to learning discriminative representations for unlabeled target instances by aligning the conditional distributions of source and target domains. However, we do not have label information for unlabeled target instances. One feasible way is to utilize pseudo-labels~\cite{GJDA,CDLS} predicted by the shared classifier directly, but there is no doubt that the adaptation capability would be impeded by falsely-assigned pseudo-labels, which will cause the accumulation of negative effect during the alignment process. To circumvent the uncertainty of pseudo-labels, we resort to uncovering the intrinsic geometric knowledge underlying data. Thus we design a geometric pseudo-label refinement mechanism to assist in assigning pseudo-labels for those instances that present geometric similarity to the category centroids of supervised data in the feature space. We first compute the centroid $\boldsymbol{\mu}^{(k)} \in \R^{d_{common}}$ of category $k$ in the labeled source and target domains, which is a mean vector of both encoded features in each category and is calculated by: 
\begin{small}
    \begin{align}\label{eq-GS-centroid}
        \boldsymbol{\mu}^{(k)} = \frac{1}{\lvert \X_S^{(k)} \cup \X_L^{(k)} \rvert} \left(\sum_{\x_{i} \in \X_S^{(k)}} \f_S(\x_{i})+\sum_{\x_{j}\in\X_L^{(k)}}\f_T(\x_{j})\right) \,,
    \end{align}
\end{small}where $\X_L^{(k)}$ denotes the set of labeled target instances of class $k$. Then, a set of centroids $\{\boldsymbol{\mu}^{(k)}\}_{k=1}^K$ are obtained and we use a geometric similarity metric to assign $i$-th unlabeled target instance with a geometric similarity label:
\begin{small}
    \begin{align}\label{eq-GS-label}
        y_{u_i}^{<GS>} = \arg\max_{k} GS\left(\f_T(\x_{u_i}), \boldsymbol{\mu}^{(k)}\right) \,,
    \end{align}
\end{small}where $GS(\cdot, \cdot)$ is the geometric relationship between two data points in the latent feature space
 and the Cosine similarity is chosen in this paper. 
Moreover, 
it is easy to get pseudo (predicted) label from the shared classifier $C$ and we define as $y_{u_i}^{<NN>}$. With geometric similarity label and neural network label, an unlabeled target instance will only be selected and assigned with pseudo-label if $y_{u_i}^{<GS>} = y_{u_i}^{<NN>}$, which can boost pseudo-label assignment accuracy.
Thus, we have $\X_T = \X_L\cup\hat{\X}_U$ and $\Y_T = \Y_L\cup\hat{\Y}_U$, where $\hat{\X}_U\,,\hat{\Y}_U$ are the sets of selected unlabeled target instances and their corresponding pseudo-labels. 

Intuitively, few unlabeled target samples are likely to be correctly annotated in the early training phase. As the training evolves, more and more samples will potentially be assigned with an agreed pseudo-label. Ultimately, the majority of samples will be assigned with confident pseudo-labels. 
Note that this automatic mechanism does not require manual experience to set the threshold. 
This hard example mining process progressively encourages those instances with consistent predicted label to be accepted to participate in the explicit semantic alignment while filtering out instances without consistency.
Accordingly, we are able to adequately learn semantic representations for unlabeled target instances using the designed explicit semantic alignment loss. Formally, there are three centroids of source, target, and combination of source and target in the common feature space for each class, constituting the triplet centroids: 
\begin{small}
    \begin{align}\label{eq-triplet-centroids}
        \boldsymbol{\mu}_S^{(k)}&=\frac{1}{\lvert \X_S^{(k)} \rvert}\sum_{\x_{i}\in\X_S^{(k)}}\f_S(\x_{i}) \,,
        \boldsymbol{\mu}_T^{(k)}=\frac{1}{\lvert \X_T^{(k)} \rvert}\sum_{\x_{j}\in\X_T^{(k)}}\f_T(\x_{j}) \,, \notag \\
        \boldsymbol{\mu}_{ST}^{(k)}&=\frac{1}{\lvert \X_S^{(k)} \cup \X_T^{(k)} \rvert} \left(\sum_{\x_{i} \in \X_S^{(k)}} \f_S(\x_{i})+\sum_{\x_{j}\in\X_T^{(k)}}\f_T(\x_{j})\right) \,,
    \end{align}
\end{small}where $\X_T^{(k)}$ represents the subset of $\X_T$ including labeled target instances whose ground-truth labels are class $k$ and unlabeled target instances with consentaneously assigned pseudo-labels of class $k$.

Once the centroids are calculated, they can facilitate semantic alignment and enhance better semantic consistency between domains. Drawing inspiration from the maximum mean discrepancy (MMD) technique which has been proved to be an effective measurement of the divergence between two domains~\cite{MMD,DDC,TPN}. Naturally, we employ the following explicit semantic alignment loss to learn more robust and discriminative representations between domains: 
\begin{small}
    \begin{align}\label{eq-ESA-loss}
        \mathcal{L}_{ESA} = \sum_{k=1}^K \left(\left \| \boldsymbol{\mu}_S^{(k)} - \boldsymbol{\mu}_T^{(k)}  \right \|^2_2 + \left \| \boldsymbol{\mu}_S^{(k)} - \boldsymbol{\mu}_{ST}^{(k)}  \right \|^2_2 + \left \| \boldsymbol{\mu}_T^{(k)} - \boldsymbol{\mu}_{ST}^{(k)}  \right \|^2_2 \right) \,.
    \end{align}
\end{small}By minimizing this objective, the centroids of each category will be forced to be in close proximity in the encoded feature subspace, resulting in semantically consistent representations across domains. 

Furthermore, the explicit semantic alignment loss enforces semantic consistency in local distributions, while ignoring the global distributions between domains. To this end, we additionally add a domain discriminator $D$ with a single layer to distinguish whether the encoded features are from source or target domain while encoders are trained to fool $D$~\cite{DANN_JMLR}. In other words, $D$ simply performs the binary classification task. Concretely,
\begin{small}
    \begin{align}\label{eq-D-supervised-loss}
        \mathcal{L}_{D} &= \frac{1}{n_s}\sum_{\x_i \in \X_S}\log\left(D(\f_S(\x_i))\right) \notag \\
        & + \frac{1}{n_l+n_u}\sum_{\x_j \in \X_L \cup \X_U}\log\left(1-D(\f_T(\x_j))\right)\,.
    \end{align}
\end{small}The feature representations from encoders $E_S\,,E_T$ are domain-invariant when this minimax game reaches an equilibrium.

\subsection{Overall Formulation and Optimization}

In summary, we exploit the implicit semantic knowledge in the prediction space for the preservation of category correlations and align features of both domains in each category to facilitate better alignment and transferability. To preserve semantic correlations and keep semantic consistency between domains, we design an implicit semantic correlation loss $\mathcal{L}_{ISC}$ to force the target network to learn the correlations of source predictions, and an explicit semantic alignment loss $\mathcal{L}_{ESA}$ with the help of geometric pseudo-label refinement mechanism to match the centroids of source, target and source-target combined features for each category. In addition, a domain confusion loss $\mathcal{L}_{D}$ is introduced to reduce the global domain divergence for improving stability. To this end, we present our overall objective function in a minimax scheme as:
\begin{small}
    \begin{align}\label{eq-overall-loss}
        \min_{C\,,E_S\,,E_T}\max_{D} \mathcal{L}_{S} + \mathcal{L}_{ISC} + \beta \mathcal{L}_{ESA} + \gamma \mathcal{L}_{D}\,,
\end{align}
\end{small}where the hyper-parameters $\beta$ and $\gamma$ balance the influences of $\mathcal{L}_{ESA}$ and $\mathcal{L}_{D}$ on the optimization process, respectively.


\begin{table*}[htbp]
  \centering
  \caption{Classification results (\%) with standard deviations for text-to-image transfer scenario.}\vspace{-4mm}
    \setlength{\tabcolsep}{0.8mm}{
    \begin{tabular}{c|ccccccccc|c}
    \toprule
    S $\rightarrow$ T & SVMt & NNt & MMDT~\cite{MMDT_ICLR} & SHFA~\cite{SHFA} & SGW~\cite{SGW} & CDLS~\cite{CDLS} & G-JDA~\cite{GJDA} & TNT~\cite{TNT} & STN~\cite{STN} & \textbf{SSAN} \\
  \hline
   N $\rightarrow$ I & 66.85$\pm$0.96 & 67.68$\pm$0.80 & 53.21$\pm$0.69 & 64.06$\pm$0.61 & 68.01$\pm$0.80 & 70.96$\pm$0.83 & 75.76$\pm$0.65 & 77.71$\pm$0.57 & 78.46$\pm$0.58 & \textbf{80.22$\pm$0.91}    \\
   \bottomrule
  \end{tabular}
    }
  \label{tab:t2i}\vspace{-2mm}
\end{table*}

\begin{table*}[htbp]
  \centering
  \caption{Classification results (\%) with standard deviations for cross-feature object recognition.}\vspace{-4mm}
    \setlength{\tabcolsep}{0.8mm}{
    \begin{tabular}{c|ccccccccc|c}
    \toprule
    S $\rightarrow$ T & SVMt & NNt & DAMA~\cite{DAMA} & MMDT~\cite{MMDT_ICLR} & SHFR~\cite{SHFR} & SHFA~\cite{SHFA} & G-JDA~\cite{GJDA} & CDLS~\cite{CDLS} & STN~\cite{STN} & \textbf{SSAN} \\
  \hline    
  \multicolumn{11}{c}{SURF to DeCAF$_6$} \\ 
  \hline
   A $\rightarrow$ A & 88.66$\pm$0.50 & 90.00$\pm$0.33 & 87.40$\pm$0.50 & 89.30$\pm$0.40 & 87.10$\pm$1.20 & 88.60$\pm$0.30 & 92.30$\pm$0.20 & 91.70$\pm$0.20 & 92.19$\pm$0.83 & \textbf{92.45$\pm$0.52}  \\
   C $\rightarrow$ C & 77.31$\pm$1.10 & 79.56$\pm$0.50 & 73.80$\pm$1.20 & 80.30$\pm$1.20 & 73.20$\pm$1.60 & 78.20$\pm$1.00 & 86.70$\pm$0.50 & 81.80$\pm$1.10 & 82.92$\pm$0.92 & \textbf{87.01$\pm$0.71}  \\
   W $\rightarrow$ W & 89.32$\pm$1.10 & 91.42$\pm$0.73 & 87.20$\pm$0.70 & 87.30$\pm$0.80 & 87.50$\pm$1.30 & 90.00$\pm$1.00 & 89.40$\pm$0.90 & 95.20$\pm$0.90 & 95.43$\pm$0.78 & \textbf{96.66$\pm$0.62}  \\
   \hline
   \multicolumn{11}{c}{DeCAF$_6$ to SURF} \\ 
   \hline
   A $\rightarrow$ A & 43.03$\pm$0.90 & 42.82$\pm$1.77 & 38.10$\pm$1.10 & 40.50$\pm$1.30 & 44.50$\pm$1.10 & 42.90$\pm$1.00 & 50.30$\pm$0.70 & 46.40$\pm$1.00 & 47.62$\pm$1.49 & \textbf{52.91$\pm$0.96} \\
   C $\rightarrow$ C & 30.15$\pm$1.50 & 31.33$\pm$2.89 & 18.90$\pm$1.30 & 30.60$\pm$1.70 & 33.40$\pm$1.00 & 29.40$\pm$1.50 & 33.70$\pm$0.80 & 31.80$\pm$1.20 & 30.83$\pm$2.14 & \textbf{37.24$\pm$0.95} \\
   W $\rightarrow$ W & 55.28$\pm$1.00 & 60.87$\pm$1.58 & 47.40$\pm$2.10 & 59.10$\pm$1.20 & 54.30$\pm$0.90 & 62.20$\pm$0.70 & 63.80$\pm$0.90 & 63.10$\pm$1.10 & 64.71$\pm$1.62& \textbf{69.81$\pm$0.88} \\
   \bottomrule
  \end{tabular}
  }
  \label{tab:i2i-cross-feature}\vspace{-2mm}
\end{table*}

\begin{table*}[htbp]
  \centering
  \caption{Classification results (\%) with standard deviations for cross-domain object recognition (SURF to DeCAF$_6$).}\vspace{-4mm}
    \setlength{\tabcolsep}{0.8mm}{
    \begin{tabular}{c|ccccccccc|c}
    \toprule
    S $\rightarrow$ T & SVMt & NNt & MMDT~\cite{MMDT_ICLR} & SHFA~\cite{SHFA} & CDLS~\cite{CDLS} & SGW~\cite{SGW} & G-JDA~\cite{GJDA} & TNT~\cite{TNT} & STN~\cite{STN} & \textbf{SSAN} \\
  \hline
   A $\rightarrow$ C & 79.64$\pm$0.46  & 81.03$\pm$0.50 & 75.62$\pm$0.57  & 71.16$\pm$0.73 & 78.73$\pm$0.49  & 79.88$\pm$0.53 & 86.60$\pm$0.17 & 85.79$\pm$0.42 & 88.21$\pm$0.16  & \textbf{88.36$\pm$0.73}   \\
   A $\rightarrow$ D & 92.60$\pm$0.71 & 92.99$\pm$0.63 & 91.65$\pm$0.83 & 95.16$\pm$0.36 & 94.45$\pm$0.59 & 93.43$\pm$0.67 & 90.67$\pm$0.65 & 92.04$\pm$0.76 & 96.42$\pm$0.43 & \textbf{97.00$\pm$0.77} \\
   A $\rightarrow$ W & 89.34$\pm$0.94 & 91.13$\pm$0.73  & 89.28$\pm$0.77 & 88.11$\pm$1.01 & 91.57$\pm$0.81 & 90.26$\pm$0.84 & 94.09$\pm$0.67 & 91.26$\pm$0.72 & \textbf{96.68$\pm$0.43} &  96.31$\pm$0.64\\
   \hline
   C $\rightarrow$ A & 89.13$\pm$0.39 & 89.60$\pm$0.33 & 87.06$\pm$0.47 & 85.49$\pm$0.51 & 86.34$\pm$0.74 & 89.03$\pm$0.37 & 92.49$\pm$0.12 & 92.35$\pm$0.17 & 93.03$\pm$0.16 & \textbf{93.16$\pm$0.51} \\
   C $\rightarrow$ D & 92.60$\pm$0.71 & 92.99$\pm$0.63 & 91.46$\pm$0.85 & 94.25$\pm$0.50 & 90.43$\pm$0.79 & 93.43$\pm$0.67 & 88.62$\pm$0.76 & 92.67$\pm$0.80 & 96.06$\pm$0.50 &  \textbf{97.44$\pm$0.61}\\
   C $\rightarrow$ W & 89.34$\pm$0.94 & 91.13$\pm$0.73 & 89.11$\pm$0.76 & 89.47$\pm$0.90 & 88.60$\pm$0.80 & 90.26$\pm$0.84 & 92.64$\pm$0.54 & 92.98$\pm$0.75 & \textbf{96.38$\pm$0.38} & 95.87$\pm$0.72\\
   \hline 
   W $\rightarrow$ A & 89.13$\pm$0.39 & 89.60$\pm$0.33 & 87.00$\pm$0.47 & 88.83$\pm$0.45 & 87.51$\pm$0.44 & 89.02$\pm$0.37 & 92.28$\pm$0.15 & 92.99$\pm$0.14 & 93.11$\pm$0.16 & \textbf{93.54$\pm$0.44} \\
   W $\rightarrow$ C & 79.64$\pm$0.46 & 81.03$\pm$0.50 & 75.44$\pm$0.59 & 79.66$\pm$0.52 & 77.30$\pm$0.71 & 79.85$\pm$0.53 & 84.82$\pm$0.38 & 86.28$\pm$0.51 & 87.22$\pm$0.45 & \textbf{88.18$\pm$0.61}  \\
   W $\rightarrow$ D & 92.60$\pm$0.71 & 92.99$\pm$0.63 & 91.77$\pm$0.83 & 95.31$\pm$0.63 & 92.72$\pm$0.75 & 93.43$\pm$0.67 & 95.87$\pm$0.41 & 94.09$\pm$0.88 & 96.38$\pm$0.57 & \textbf{97.64$\pm$0.51} \\ 
   \hline
   Avg. & 87.68$\pm$0.63 & 88.69$\pm$0.55 & 86.49$\pm$0.68 & 87.49$\pm$0.62 & 87.52$\pm$0.68 & 88.73$\pm$0.61 & 90.90$\pm$0.43 & 91.17$\pm$0.44 & 93.72$\pm$0.36 & \textbf{94.17$\pm$0.97}\\
   \bottomrule
  \end{tabular}
    }
  \label{tab:i2i_surf2decaf}\vspace{-2.5mm}
\end{table*}

\begin{table}[!htbp]
  \centering
  \caption{Classification results (\%) with standard deviations for cross-domain object recognition (SURF to ResNet50).}\vspace{-4mm}
  \setlength{\tabcolsep}{0.8mm}{
  \begin{tabular}{c|cccc|c}
    \toprule
    S $\rightarrow$ T & SVMt & NNt & CDLS~\cite{CDLS} & STN~\cite{STN} & \textbf{SSAN} \\
  \hline
   A$\rightarrow$C & 19.99$\pm$1.78 & 20.22$\pm$1.96 & 21.32$\pm$2.01 & 21.64$\pm$2.36 & \textbf{24.07$\pm$2.00}   \\
   A$\rightarrow$D & 84.37$\pm$3.54 & 84.25$\pm$3.78 & 85.03$\pm$1.97 & 85.20$\pm$2.80 & \textbf{91.81$\pm$3.19} \\
   A$\rightarrow$W & 78.60$\pm$3.36 & 81.51$\pm$3.01 & 80.38$\pm$3.02 & 83.77$\pm$3.19 & \textbf{88.47$\pm$1.21} \\
   \hline
   C$\rightarrow$A & 34.26$\pm$3.13 & 34.81$\pm$3.51 &34.69$\pm$1.83 & 34.16$\pm$3.11 & \textbf{41.71$\pm$2.07} \\
   C$\rightarrow$D & 84.37$\pm$3.54 & 84.25$\pm$3.78 & 86.61$\pm$2.76 & 86.59$\pm$3.57 & \textbf{91.22$\pm$2.35} \\
   C$\rightarrow$W & 78.60$\pm$3.36 & 81.51$\pm$3.01 & 78.11$\pm$2.08 & 84.21$\pm$2.56 & \textbf{86.24$\pm$2.04} \\
   \hline 
   W$\rightarrow$A & 34.26$\pm$3.13 & 34.81$\pm$3.51 & 35.66$\pm$1.19 & 35.18$\pm$2.79 & \textbf{37.36$\pm$1.98} \\
   W$\rightarrow$C & 19.99$\pm$1.78 & 20.22$\pm$1.96 & 20.31$\pm$1.01 & 20.71$\pm$2.66 & \textbf{23.50$\pm$2.19} \\
   W$\rightarrow$D & 84.37$\pm$3.54 & 84.25$\pm$3.78 & 82.68$\pm$1.57 & 83.74$\pm$2.96 & \textbf{93.62$\pm$2.27} \\
   \hline
   Avg. & 57.65$\pm$3.02 & 58.43$\pm$3.14 & 58.31$\pm$1.94 & 59.47$\pm$2.89  & \textbf{64.22$\pm$2.14} \\
   \bottomrule
  \end{tabular}
  }
  \label{tab:i2i_surf2resnet50}\vspace{-4mm}
\end{table}

\section{Experiment}


\subsection{Datasets and Setup}

\textbf{NUS-WIDE}~\cite{NUS-WIDE} and \textbf{ImageNet}~\cite{imagenet_cvpr09} are used for text-to-image transfer task. The former contains 269,648 images with tag information from Flickr, while the latter includes 5,247 synsets and 3.2 million images. We use the tag information of NUS-WIDE (N) and the image data of ImageNet (I) as the source domain (text) and the target domain (image), respectively. The lack of co-occurrence between text and image domain data makes target image classification a challenging cross-modal learning task. Based on the protocol in~\cite{TNT,STN}, 8 overlapping categories of these two datasets are chosen. We pre-process the NUS-WIDE tag data by pre-training a 5-layer neural network with a softmax layer and extract the 4-th hidden layer as the 64-dimensional features for text data. We then follow~\cite{tommasi2014a} to extract 4096-dimensional DeCAF$_6$ features for image data. For the source domain, 100 texts per category in NUS-WIDE are chosen to form the labeled source data. As for the target domain (ImageNet), 3 images in each class are randomly sampled to be used as labeled target data and all remaining images are used for testing. 

\textbf{Office+Caltech-256}~\cite{Office31,caltech}. The former is composed of 4,652 images belonging to 31 classes collected from three domains: Amazon (A), Webcam (W) and DSLR (D). The latter includes 30,607 images with 256 object classes from Caltech-256 (C). 10 overlapping classes are chosen to construct the Office+Caltech-256 dataset for image-to-image experiments~\cite{HFA,CDLS}. Furthermore, we consider three types of feature representations: 800-dimensional SURF features~\cite{SURF} and 4096-dimensional DeCAF$_6$ features~\cite{DeCAF}, as well as 2,048-dimensional ResNet50 features~\cite{resnet}. Among these transfer tasks, all source domain images are utilized. As suggested by~\cite{GJDA,CDLS,TNT,STN}, for the target domain, we randomly choose 3 images per class as the labeled target data, while all remaining images are set as unlabeled target data to be recognized. Note that DSLR is only viewed as the target domain due to the limited amount of images. To testify the effectiveness and robustness of our SSAN model against various HDA scenarios, two kinds of transfer tasks are conducted: (1) Transfer tasks across heterogeneous features within the same domain (i.e., A $\rightarrow$ A, C $\rightarrow$ C, W $\rightarrow$ W), where SURF $\rightarrow$ DeCAF$_6$ and DeCAF$_6$ $\rightarrow$ SURF are two settings adopted for source and target features, respectively. (2) Transfer tasks across heterogeneous features and distinct domains, in which SURF features are used as features for the source domain and DeCAF$_6$ or ResNet50 features are used as features for the target domain. Also, source and target data come from distinct domains. We evaluate our method on 9 adaptation scenarios: A $\rightarrow$ C, $\ldots$, W $\rightarrow$ D. By performing all these tasks, we can verify that our SSAN is effective and robust to tackle heterogeneities including diverse features and distinct domains. 

\textbf{Multilingual Reuters Collection}~\cite{text_dataset}. This multilingual text categorization dataset consists of about 11,000 articles from 6 categories written in five different languages: English (EN), French (FR), German (GR), Italian (IT) and Spanish (SP). Following~\cite{GJDA,CDLS,HFA,SHFA}, all the articles are represented by Bag-of-Words (BOW) with TF-IDF features, and then perform dimensionality reduction of features using PCA with 60\% energy preserved. The reduced dimensions of five languages EN, FR, GE, IT and SP are 1,131, 1,230, 1,471, 1,041 and 807, respectively. We consider EN, FR, GR and IT as the source domain and SP as the target one, following \cite{CDLS,STN}. For the source domain, 100 articles per category are randomly selected to constitute the labeled source data. For the target domain, we randomly pick up $m$ (i.e., $m = 5, 10, 15, 20$) and 500 articles per category as labeled and unlabeled target data, respectively. By choosing different amounts of labeled data, we can investigate how the number of labeled target data affects the transferability. 

\textbf{Implementation Details.} We implement our SSAN using the PyTorch framework~\cite{paszke2019pytorch}. Both the source and target feature encoders are two-layer neural networks, which apply Leaky ReLU~\cite{leakyrelu} as the activation function following~\cite{STN}. The classifier and the domain discriminator are both neural networks with a single layer. As for parameter settings, we empirically set $\alpha$ = 0.1, $\beta$ = 0.004, $\gamma$ = 0.01, $T = 5$ and the dimension of the common subspace $d_{common}$ is set to 256. 
Parameter sensitivity is conducted in Section \ref{sec:Analyses} to testify the robust performance on a wide range of parameter values. 

\vspace{-3mm}
\subsection{Evaluation}
We evaluate the performance of the SSAN by first considering SVMt and NNt as baseline methods, which simply train a support vector machine and a two-layer neural network with only the labeled target data, respectively. Then, we compare our method against several state-of-the-art HDA methods (e.g., MMDT~\cite{MMDT_ICLR}, CDLS~\cite{CDLS}, TNT~\cite{TNT}, STN~\cite{STN} etc.)
 to demonstrate the potency of our method. We apply the classification accuracy of unlabeled target data as the evaluation metric, which is widely used in existing literature~\cite{CDLS,SHFA,STN}. 
Note that partial reported results are copied from their published papers if the experimental setup is the same. 

\textbf{Text-to-Image transfer task:} 
Table~\ref{tab:t2i} presents the results of transfer task from NUS-WIDE to ImageNet. When performing alignment and classification on this task, our method achieves the best performance among all methods. Compared with the best supervised method NNt, our SSAN achieves the performance boost of more than \textbf{12\%}. In addition, our method also outperforms STN, the best-performed baseline, by a large margin. Due to the presence of challenging heterogeneities between this text to image adaptation task, even though the soft-label strategy is utilized in STN, it is still not enough to avert the negative effects of false pseudo-labels. Thus, it would be desirable to consider both semantic correlations between classes and progressively conduct semantic alignment. The performance gains achieved by our SSAN over other state-of-the-art HDA methods could indicate the effectiveness of SSAN.

\begin{figure*}[t]
  \centering
  \includegraphics[width=0.98\textwidth]{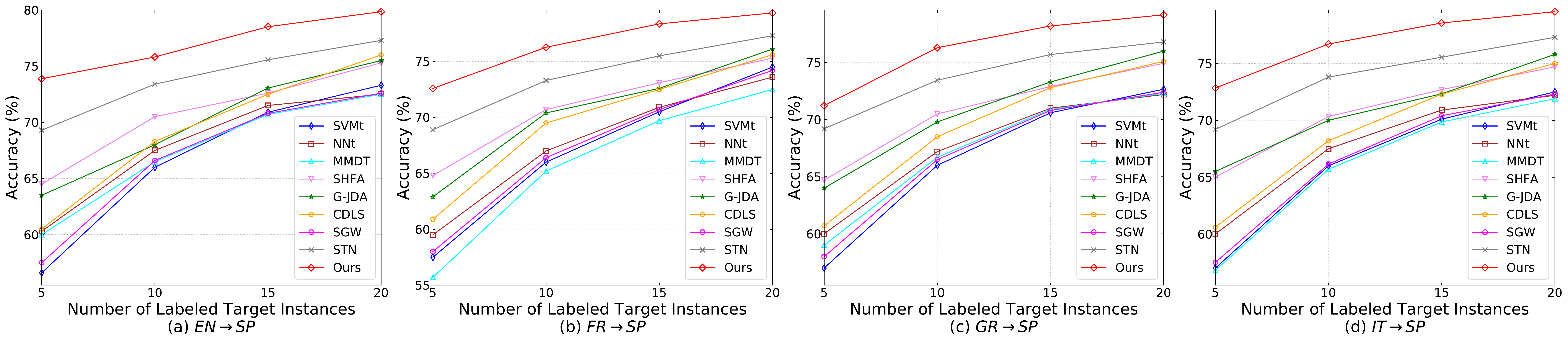}\vspace{-4mm}
  \caption{Classification results w.r.t. different number of labeled target instances per class on the multilingual Reuters dataset.}
  \label{Fig_text}\vspace{-3mm}
  \end{figure*}
  
\textbf{Image-to-Image transfer tasks:}
The classification results of transfer tasks across features within the same domain are summarized in Table~\ref{tab:i2i-cross-feature}. Based on the experimental results from Table~\ref{tab:i2i-cross-feature}, it is desirable that our SSAN obtains significant improvements over the previous methods on all six tasks. The convincing results successfully verify that our model could effectively tackle features where heterogeneities present and is robust regarding domains. On the other hand, the classification results of transfer tasks across features and domains are shown in Table~\ref{tab:i2i_surf2decaf} and Table~\ref{tab:i2i_surf2resnet50}. In terms of transferring between shallow features (SURF) and diverse deep features (DECAF$_6$ and ResNet50), our SSAN reaches the highest average accuracies of \textbf{94.17}\% and \textbf{64.22\%} among all the compared methods, indicating the robustness and effectiveness of SSAN when coping with diverse feature spaces. Ultimately, the above results support that our SSAN could capture enriched semantic information to facilitate learning more transferable feature representations. 

\textbf{Text-to-Text transfer task:}
In Figure~\ref{Fig_text}, with respect to text-to-text transfer task, we compare SSAN with previous approaches based on the different number of labeled target instances per class. Our model outperforms the comparisons among all the tasks. For the most challenging tasks with only 5 labeled target instances being available, the average classification accuracy over all these four tasks (i.e., EN$\rightarrow$SP$_5$, FR$\rightarrow$SP$_5$, GE$\rightarrow$SP$_5$, IT$\rightarrow$SP$_5$) is \textbf{72.53\%}, which outperforms the best HDA method, i.e., STN, by \textbf{2.68\%}. Intuitively, we could observe that by increasing the number of labeled target instances, better alignment is facilitated and the performance of all methods on all tasks is boosted. It is also reasonable to see that most HDA methods can yield either comparable or superior performance compared with supervised methods, i.e., NNt and SVMt, which demonstrates that HDA methods can facilitate divergence minimization and positive transfer in all text-to-text transfer tasks.

\begin{table}[!htbp]
  \centering
  \caption{Ablation study of SSAN.}\vspace{-4mm}
  \begin{tabular}{cccc}
    \toprule
  Method/Task & N $\rightarrow$ I & C $\rightarrow$ D & EN $\rightarrow$ SP$_5$ \\
  \midrule
  SSAN ($\alpha=0$) & 73.24$\pm$1.15 & 96.22$\pm$0.91 & 66.91$\pm$0.89  \\
  SSAN ($\beta=0$) & 69.21$\pm$1.80 & 96.09$\pm$0.78 & 64.39$\pm$1.96  \\
  SSAN ($\gamma=0$) & 77.59$\pm$1.65 & 96.30$\pm$0.86 & 71.98$\pm$1.01  \\
  SSAN (w/o $T$)   & 71.34$\pm$0.80 & 96.14$\pm$1.07 & 65.48$\pm$2.15  \\
  SSAN (w/o $GS$) & 77.30$\pm$0.98 & 96.30$\pm$0.94 & 69.11$\pm$1.54 \\
  \midrule
  \textbf{SSAN (full)} & \textbf{80.22$\pm$0.91} & \textbf{97.44$\pm$0.61} & \textbf{73.68$\pm$0.50} \\
  \bottomrule
  \end{tabular}
  \label{tab:ablation}\vspace{-4mm}
\end{table}

\subsection{Insight Analyses}
\label{sec:Analyses}
\textbf{Ablation Study.} Ablation study is conducted by evaluating several variants of SSAN: (1) SSAN ($\alpha=0$), which removes $\mathcal{L}_{soft}$ from Eq. (\ref{eq-ISC-loss}); (2) SSAN ($\beta=0$), which removes the explicit semantic alignment; (3) SSAN ($\gamma=0$), which turns off the $\mathcal{L}_D$; 
(4) SSAN (w/o $T$), which removes the temperature from our model (i.e., $T=1$ in Eq. (\ref{eq-source-soft})); (5) SSAN (w/o $GS$), which assigns pseudo-labels for unlabeled target instances only by the neural network (shared classifier) predictions instead of the integration of predictions from both the neural network and the geometric similarity mechanism. 

As shown in Table~\ref{tab:ablation}, the results are performed on randomly choosing N $\rightarrow$ I, C (SURF) $\rightarrow$ D (DeCAF$_6$), and EN $\rightarrow$ SP$_5$ as the representative tasks for text-to-image, image-to-image, and text-to-text transfer scenarios. We can clearly observe that the full SSAN outperforms all its variants by a large margin, which indicates that any one of these components plays an indispensable role and brings benefits to facilitate the positive transfer. Among all these components, implicit semantic correlation knowledge transfer and explicit semantic alignment bring the top two highest total gain of three tasks, which manifests the importance of transferring both explicit and implicit semantic knowledge. On the other hand, the performance gain contributed by the domain adversarial alignment in SSAN ($\gamma=0$) is inferior to the semantic alignment. This observation suggests that despite the positive effect of domain  alignment, it possesses less importance compared with semantic alignment components. With regard to SSAN (w/o $T$), it further indicates that higher temperature $T$ would have a positive performance gain since setting $T=1$ could not fully explore the correlations among categories during semantic transfer. 
\begin{figure}[!htbp]
  \centering
  \includegraphics[width=0.48\textwidth]{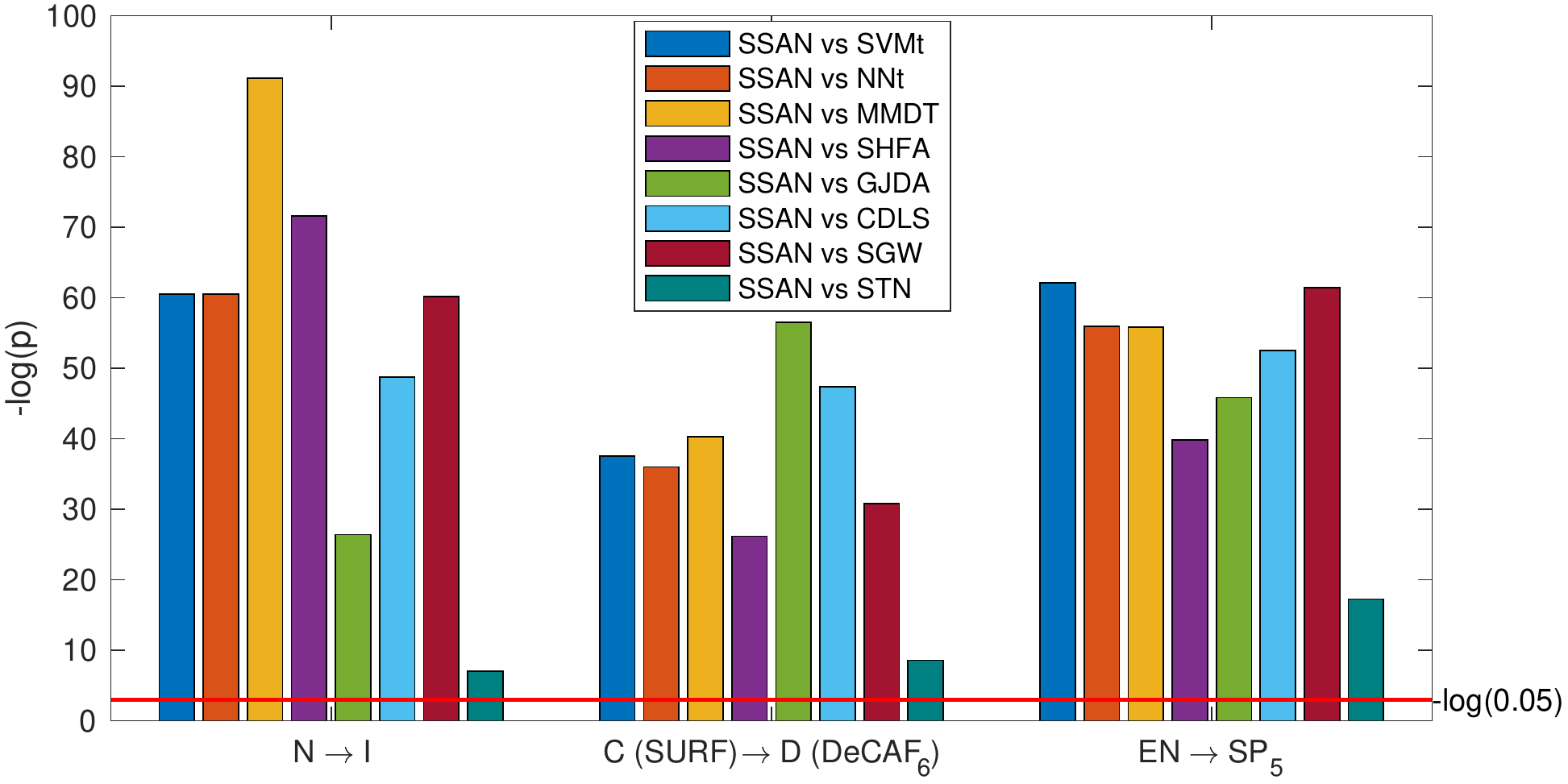}\vspace{-4mm}
  \caption{The significant test with 0.05 as the significance level is conducted to testify whether the performance gains of our SSAN compared with other baselines on three transfer tasks are statistically significant. The larger the value is, the more statistically significant the result indicates. }
  \label{Fig_pvalue}
\end{figure}
Furthermore, the inclusion of geometric pseudo-label refinement enhances the performance by \textbf{2.88\%} on average, which verifies that the geometric pseudo-label refinement can facilitate a more accurate pseudo-label assignment and boost the performance. 

\begin{figure*}[!htbp]
  \centering
  \includegraphics[width=0.98\textwidth]{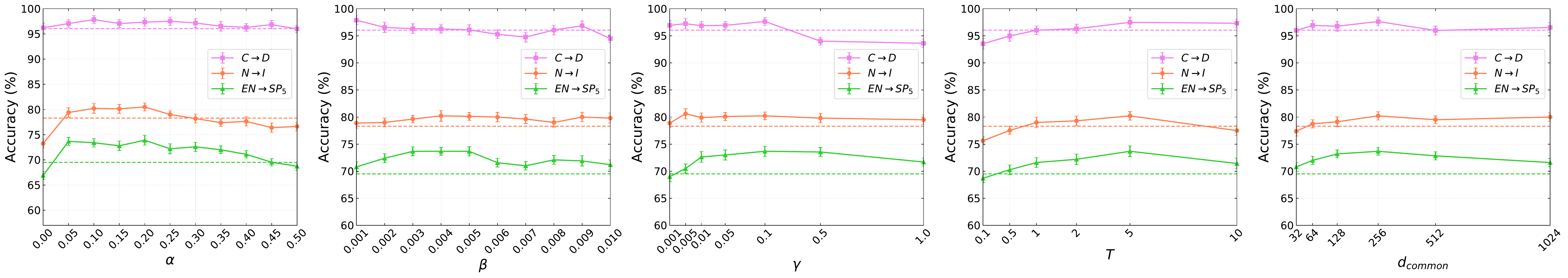}\vspace{-4mm}
  \caption{Parameter sensitivity on three random transfer tasks. The dashed lines show the corresponding best baseline results.}\vspace{-2mm}
  \label{Fig_parameter}
\end{figure*}

\textbf{Significance Test (t-Test).}
To further validate the superiority of the SSAN, a statistical significance test is shown in Figure~\ref{Fig_pvalue}. A significant level of 0.05 is applied, and a $p$-value that is less than 0.05 indicates that the performance boost of our SSAN compared with another method is statistically significant. In Figure~\ref{Fig_pvalue}, we illustrate the $-\log(p)$ for each $p$-value, and a red line which indicates the base significance level of 0.05 ($-\log(0.05)$). The larger the value of $-\log(p)$ is, the more statistically significant the result indicates. Based on the results, it verifies the significant superiority of our SSAN compared with other comparable approaches in all three distinct transfer scenarios, which further demonstrates the effectiveness and robustness of the SSAN.

\textbf{Parameter Sensitivity.}
To validate the parameter sensitivity of our model, we randomly take N $\rightarrow$ I, C (SURF) $\rightarrow$ D (DeCAF$_6$), and EN $\rightarrow$ SP$_5$ as the representative tasks. We vary parameters including $\alpha$, $\beta$, $\gamma$, $T$ and the dimension of common subspace $d_{common}$ within their corresponding reasonable ranges, and plot the results in Figure~\ref{Fig_parameter}. We can observe that the results demonstrate the robustness of SSAN under a wide range of parameter choices, which indicates that our model is stable and effective. 

\begin{figure}[!htbp]
  \centering
  \includegraphics[width=0.48\textwidth]{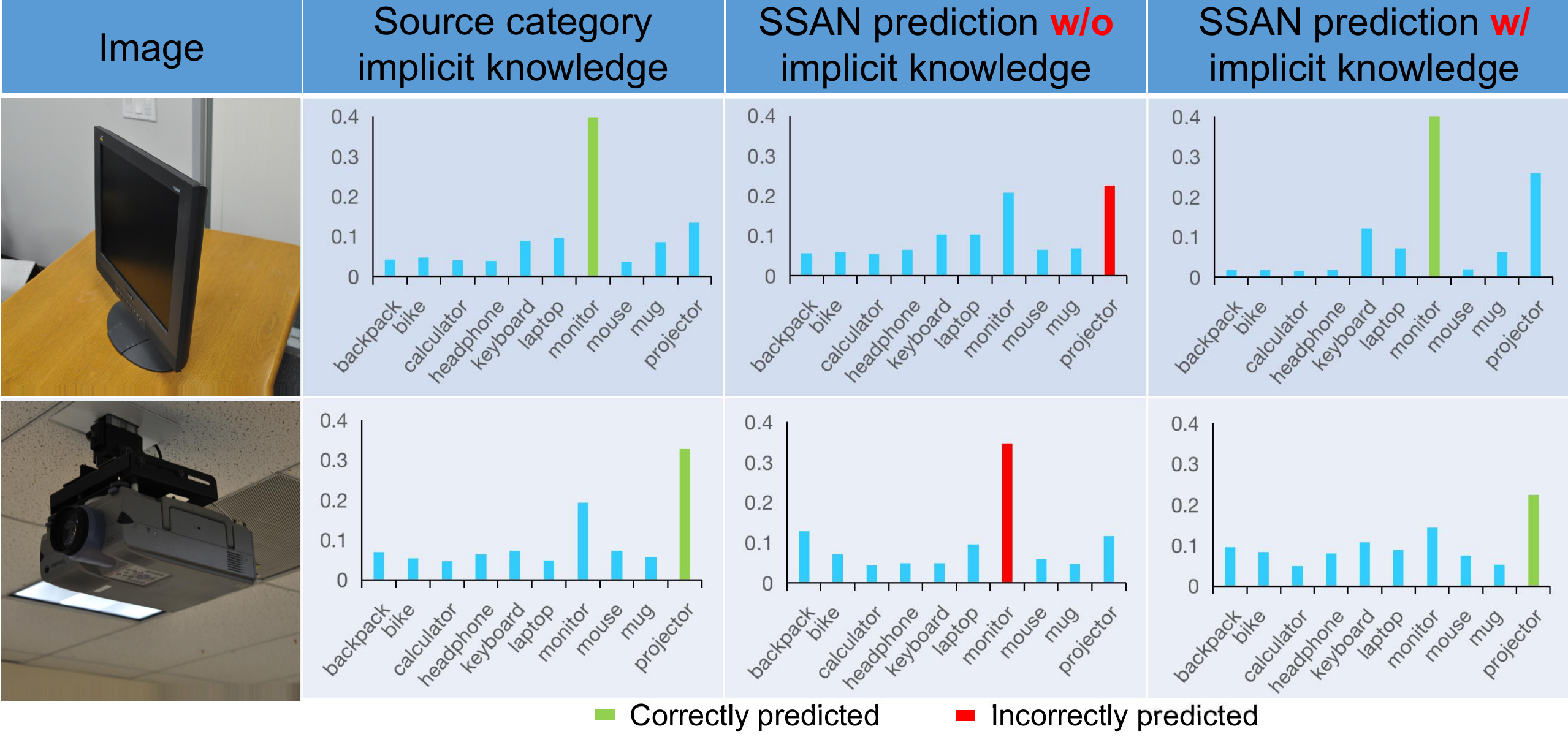}\vspace{-3mm}
  \caption{Two illustrative examples, their corresponding source semantic correlation knowledge, and predicted distributions without/with implicit knowledge are presented. }
  \label{Fig_correlation}\vspace{-4mm}
\end{figure}

\textbf{Implicit Semantic Correlation Visualization.}
We investigate the efficacy of our implicit semantic correlation knowledge and show it in Figure~\ref{Fig_correlation}. The leftmost histograms illustrate the average prediction distribution of source instances that belong to the category of the shown images. They implicitly express the semantic correlation relationships among categories. By leveraging this implicit knowledge during the alignment process, the learner can learn the semantic correlations, for instance, the target monitor category can mimic a similar predictive distributions just as what the source monitor category has. Therefore, applying this implicit knowledge can transfer the correlation relationship implicitly contained in source predictions which will facilitate the alignment process and mitigate the possibility of category mismatches. Moreover, the implicit knowledge can correct some wrong predictions and hence prevent the model from being confused and learn a more robust classifier. The middle and the rightmost histograms show the predicted distributions of the shown image instances without using or using the implicit semantic correlation transfer, respectively. The results verify that by taking implicit knowledge into account, our SSAN can successfully classify both instances.

\begin{figure}[!htbp]
  \centering
  \includegraphics[width=0.48\textwidth]{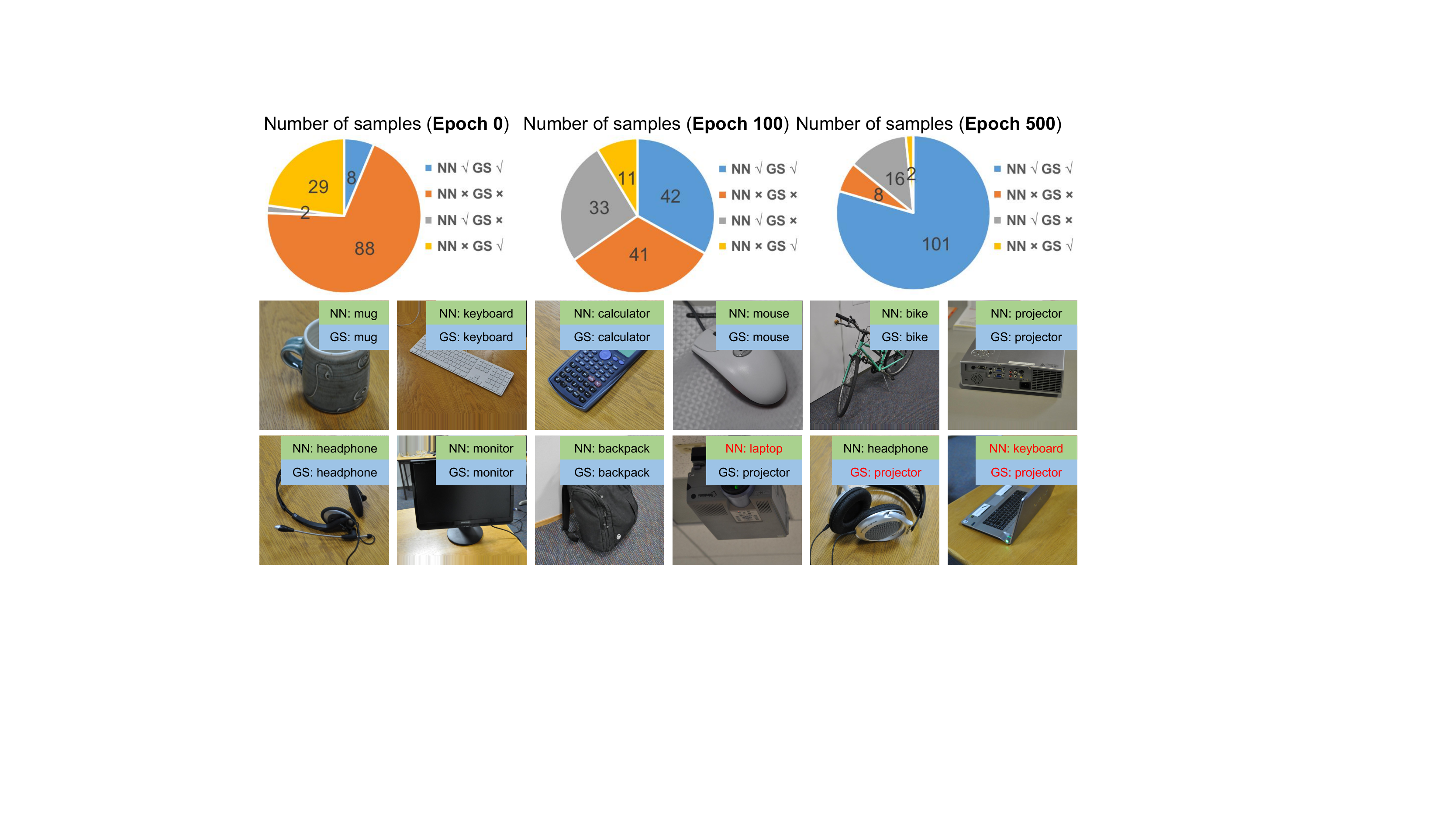}\vspace{-3mm}
  \caption{Top: Pie charts statistically summarize the amount of correctly and incorrectly assigned pseudo-labels predicted by NN (neural network) and GS (geometric similarity) during different phases of training on task C (SURF) $\rightarrow$ D (ResNet50). Tick and cross stand for correctly assigned pseudo-label and incorrectly assigned pseudo-label, respectively. Bottom: Example images with their assigned pseudo-labels at epoch 500 are presented. The green and light-blue tags represent NN and GS assigning pseudo-labels, respectively. Incorrect pseudo-label assignments are shown in red. }
  \label{Fig_consistency}\vspace{-1mm}
\end{figure}

\textbf{Geometric Similarity Pseudo-label Refinement.}
To examine the effectiveness of our geometric pseudo-label refinement mechanism, in Figure~\ref{Fig_consistency}, pie charts are utilized to statistically summarize pseudo-label predictions of task C (SURF) $\rightarrow$ D (ResNet50) for three different phases during training, and some example images with their assigned pseudo-labels at epoch 500 are presented. 
The blue portion in pie charts represents relatively easy instances which are easy to be correctly predicted by both the neural network (NN) and the geometric similarity (GS) mechanism, which form our pseudo-label refinement procedure. The instances of this portion will be leveraged during the alignment process as the pseudo-labels assigned by two learners reaching a consensus. On the other hand, instances belonging to the grey and the yellow portion will be filtered out by our refinement procedure due to the disagreement of assigned pseudo-labels by two learners. Hence, hard instances like the projector and the instances that do not present strong geometric similarity like the headphone will not be assigned with a pseudo-label used for alignment. This effectively mitigates the negative effects caused by these instances during the alignment process. For those instances in the orange portion such as the laptop example, although both learners incorrectly predict their pseudo-labels, however, there is still a chance that the pseudo-labels will not be the same, which can still pare some hard instances down from the alignment process. Unlike previous methods such as~\cite{GJDA,CDLS} which directly utilize pseudo-labels assigned by NN without any refinement, our SSAN effectively refines pseudo-labels by considering geometric semantics. This ensemble procedure will raise the pseudo-label assignment accuracy and hence facilitate better fine-level semantic alignment. 

\section{Conclusion}

In this paper, we propose a novel Simultaneous Semantic Alignment Network (SSAN) to address the HDA problems. Compared with existing HDA methods, our method simultaneously excavates both implicit and explicit semantic knowledge to facilitate alignment between cross-domain heterogeneous data. Implicit semantic knowledge facilitates achieving better preservation of categorical distributional correlations, while explicit triplet-centroid alignment procedure with the geometric semantic label refinement enforces cross-domain semantic alignment and consistency. The experimental results on a variety of cross-modal transfer tasks demonstrate the superiority of SSAN against several state-of-the-art HDA methods. 

\section*{Acknowledgements}
This work was supported by the National Natural Science Foundation of China (61902028).


\clearpage\balance
\bibliographystyle{ACM-Reference-Format}
\bibliography{Reference_mm2020}



\end{document}